\documentclass[journal,twoside,web]{ieeecolor}
\usepackage{xcolor}
\usepackage{generic}
\usepackage{cite}
\usepackage{amsmath,amssymb,amsfonts}
\usepackage{graphicx}
\usepackage{times}
\usepackage{helvet}
\usepackage{booktabs}
\usepackage{placeins}
\usepackage{afterpage}
\usepackage{colortbl}
\definecolor{linkblue}{rgb}{0.075,0.541,0.855}
\definecolor{bestcell}{RGB}{255,226,226}
\definecolor{bestink}{RGB}{176,32,40}
\definecolor{tabhead}{RGB}{214,232,248}
\makeatletter
\let\NAT@parse\undefined
\makeatother
\usepackage{hyperref}
\hypersetup{colorlinks=true,citecolor=linkblue,linkcolor=linkblue,urlcolor=linkblue}
\makeatletter
\def\@biblabel#1{{\color{linkblue}[#1]}}

\makeatother
\usepackage{textcomp}
\AtBeginDocument{\raggedbottom}
\makeatletter
\newif\if@aftersection
\def\@afterheading{%
  \@nobreakfalse
  \everypar{%
    \@aftersectionfalse
    \clubpenalty \@clubpenalty
    \everypar{}%
  }%
}
\def\section{%
  \@aftersectiontrue
  \@startsection{section}{1}{\z@}{1.6ex minus 0.3ex}%
  {0.65\baselineskip}{\color{subblue}\centering\normalsize\usefont{T1}{phv}{m}{n}}}
\def\subsection{%
  \if@aftersection
    \@aftersectionfalse
    \expandafter\@firstoftwo
  \else
    \expandafter\@secondoftwo
  \fi
  {\@startsection{subsection}{2}{\z@}{0pt}%
    {0.65\baselineskip}{\color{subblue}\normalsize\usefont{T1}{phv}{m}{it}\raggedright}}%
  {\@startsection{subsection}{2}{\z@}{1.3ex minus 0.3ex}%
    {0.65\baselineskip}{\color{subblue}\normalsize\usefont{T1}{phv}{m}{it}\raggedright}}}
\makeatother
\makeatletter
\def\ps@titlepagestyle{%
\def\@oddhead{\vbox{\hbox to\textwidth{\scriptsize\sffamily\journalname\hfil\thepage}\vskip2pt{\color{subsectioncolor}\hrule height1pt}}}%
\let\@evenhead\@oddhead
\def\@oddfoot{}\def\@evenfoot{}}
\makeatother

\def\BibTeX{{\rm B\kern-.05em{\sc i\kern-.025em b}\kern-.08em
    T\kern-.1667em\lower.7ex\hbox{E}\kern-.125emX}}
\begin{document}
\title{SparseTokenGS: Gaussian-based Sparse-View\\CBCT Reconstruction via Reusable Token Pool}

\author{Zihan Deng, Zhenzhen Gong, Yujie Liu, Jiaxin Gong, Dingkun Liu, Kexian Xie,\\
Yixing Huang, Peng Jin, Tianye Niu, Zhisheng Wang, \IEEEmembership{Member, IEEE},\\
Fenglin Liu, \IEEEmembership{Member, IEEE}, Hongming Shan, \IEEEmembership{Senior Member, IEEE}, and Lequan Yu, \IEEEmembership{Member, IEEE}
\thanks{Z. Deng, D. Liu, and L. Yu are with the School of Computing and Data Science, The University of Hong Kong, Hong Kong, China (e-mail: zihandeng@connect.hku.hk, driverliu@163.com, lqyu@hku.hk).}
\thanks{Z. Gong, Y. Liu, and F. Liu are with the Engineering Research Center of Industrial Computed Tomography Nondestructive Testing, Ministry of Education, Chongqing University, Chongqing 400044, China (e-mail: gongzhenzhen0516@163.com, liuyj@stu.cqu.edu.cn, liufl@cqu.edu.cn).}
\thanks{K. Xie is with the Department of Stomatology, Peking University Third Hospital, Beijing 100191, China (e-mail: kexianxie@bjmu.edu.cn).}
\thanks{Y. Huang is with the Institute of Medical Technology, Peking University Health Science Center, Beijing 100191, China (e-mail: huangyx@hsc.pku.edu.cn).}
\thanks{P. Jin and T. Niu are with the Shenzhen Bay Laboratory, Shenzhen 518132, China (e-mail: jinpeng@szbl.ac.cn, niuty@szbl.ac.cn).}
\thanks{J. Gong and Z. Wang are with the Jiangxi Intelligent Non-Destructive Testing Innovation Center, Nanchang 330096, China, and also with the Key Laboratory of Nondestructive Testing (Ministry of Education), Nanchang Hangkong University, Nanchang 330063, China (e-mail: gong\_jiaxin@163.com, zhisheng.wang@nchu.edu.cn).}
\thanks{H. Shan is with the Institute of Science and Technology for Brain-inspired Intelligence, Fudan University, Shanghai 200433, China (e-mail: hmshan@ieee.org).}
\thanks{This research was supported in part by the Guangdong Hong Kong Science and Technology Cooperation Funding Project under Grant ghp/318/22gd, and in part by the Jiangxi Provincial Natural Science Foundation (Youth Program, Category C) under Grant 20262BAC200480.}
\thanks{(Corresponding authors: Lequan Yu and Zhisheng Wang.)}}

\maketitle

\begin{abstract}
Sparse-view cone-beam computed tomography is widely used for fast, low-dose volumetric imaging and lesion localization in dentistry and image-guided interventions, but it introduces severe artifacts. Feed-forward Gaussian methods transfer the reconstruction cost to training and decode each coordinate from projection-aligned features and neighboring anchors. However, information from the whole scan is written only when the field is constructed and cannot be retrieved during coordinate decoding, leaving the volume inconsistent within the scan. This is a common limitation of such methods: once the field is fixed, decoding uses only local features and reconstruction quality drops substantially. We propose SparseTokenGS, a Gaussian-based method for sparse-view cone-beam computed tomography reconstruction via a reusable token pool. Anchor Token Fusion and Point-Token Attention query this shared pool at the anchors and at the query coordinates, respectively. The local Gaussian prior is retained, and the volume is reconstructed in one forward pass, without per-scan optimization or a projection-domain loss. On three public chest and dental datasets, SparseTokenGS achieves the highest reconstruction performance among the compared methods. On clinical dental scans, reconstruction from only six projections reaches 29.51 decibels in peak signal-to-noise ratio and a structural similarity of 0.842 against the full-view reference, and off-the-shelf tooth-and-pulp segmentation keeps these structures separable for root-canal localization. Experiments on volcanic rock and chicken show performance after same-object adaptation. A reusable token pool gives feed-forward reconstruction a mechanism that keeps scan-wide context available through decoding and can benefit future rapid clinical lesion localization and analysis. The code is available at: \url{github.com/FrankDengAI/SparseTokenGS}.
\end{abstract}


\section{Introduction}
\label{sec:introduction}
\IEEEPARstart{C}{one}-beam computed tomography (CBCT) provides a volume from one gantry rotation for dental imaging, image-guided intervention, and selected head-and-neck examinations~\cite{kaasalainen2021dental}. When the orbit is densely sampled, reconstruction is typically stable and sharp. Commercial protocols often acquire hundreds to thousands of projections. Sparse angular sampling is nevertheless used to localize lesions quickly, shorten reconstruction time, or reduce radiation burden. The inverse problem is then underdetermined: points in the field of view still reach the detector in every acquired view, but the angular constraints are too few, and direct Feldkamp-Davis-Kress (FDK) reconstruction leaves conspicuous streak artifacts~\cite{feldkamp1984practical}. Iterative methods, such as the simultaneous algebraic reconstruction technique (SART) and total-variation (TV) penalized variants, can reduce the data residual without a training set~\cite{andersen1984sart,sidky2008image}. In three dimensions, however, they are slow, and penalties that suppress streaks also tend to remove low-contrast texture.

\begin{figure}[!t]
\centering
\includegraphics[width=\columnwidth]{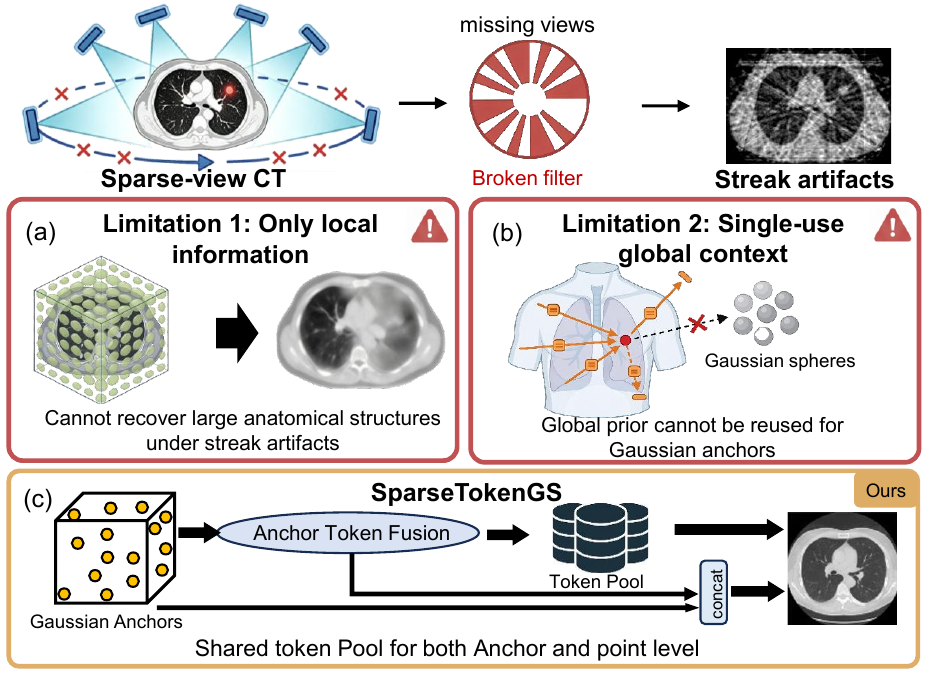}
\caption{Motivation for dual-level token conditioning in sparse-view CBCT.}
\label{fig:motivation}
\end{figure}

Deep networks are now widely used to restore corrupted FDK volumes or sparse sinograms and to predict attenuation fields directly~\cite{jin2017deep,wu2021drone,chen2018learn,adler2018primal}. Gaussian representations and neural fields are attractive for compact local modeling, efficient computation, and strong expressive power~\cite{kerbl20233d,mildenhall2021nerf}. Existing work falls into two groups. One repeatedly optimizes a density field or radiative Gaussian model on the projections of a single scan. Reconstruction quality improves with iteration, yet each patient requires a separate optimization, which is difficult to reconcile with rapid lesion localization or real-time intraoperative guidance~\cite{zha2022naf,cai2024saxnerf,zha2024r2gaussian,wu2025dgr,su2026diffnr}. To amortize reconstruction over training and improve generalization to unseen cases, the other group trains feed-forward models on large cohorts, so that one forward pass maps sparse projections to a volume~\cite{lin2024difgaussian,zhang2026xlrm,liu2025xgrm,lin2023difnet,liu2025geometry}. These methods encode multi-view evidence into local anchors, voxels, or triplanes and then decode intensity at each coordinate. As Fig.~\ref{fig:motivation} illustrates, scan-level context is written only while the field is constructed. Once that representation is fixed, coordinate decoding relies mainly on local features and cannot retrieve the appearance of the whole scan, leaving the volume prone to internal inconsistency, residual streaks, and loss of fine boundaries. Learnable tokens that predict Gaussians are a useful precedent, but those tokens emit primitives and are not retained as a conditioning pool at decoding time~\cite{ren2026tokengs}. These methods have the following property in common: once a feed-forward Gaussian field is formed, the appearance of the whole scan is no longer accessible at the coordinates being decoded. With sparse angular sampling, fine detail and consistency within the volume are then difficult to preserve together.

To address this limitation, we propose SparseTokenGS, the first feed-forward Gaussian framework that couples the context written into the anchors with the evidence retrieved at the query coordinates through one reusable token pool. A shared encoder writes a fixed-length, pose-free pool for the current scan, and the pool is read twice because the anchors and the queries need that same evidence at different stages of decoding. Anchor Token Fusion adds a zero-initialized residual to the anchor features, with geometry taken from local projection features alone, and Point-Token Attention reads the same pool at each query together with the view and neighboring Gaussian features. Volumetric supervision alone, without a projection loss or per-scan optimization, lets one forward pass return the volume.

We assess this design on simulated chest and dental cohorts and on held-out clinical dental scans. There, an off-the-shelf tooth and pulp segmenter still separates these structures, which supports lesion localization for root-canal treatment. Ablations and the volcanic rock and chicken experiments then show how the pool is used and how the model generalizes.

The main contributions of this work are summarized as follows.
\begin{enumerate}
\item We propose SparseTokenGS, the first feed-forward Gaussian framework with dual-level conditioning through one reusable, pose-free scan-level pool. One forward pass returns the volume under volumetric supervision alone.
\item We design token-conditioned Gaussian anchors and a hybrid point query. Anchor geometry comes from local projection features, the pool conditions appearance, and each query combines the view, five neighboring anchors, and a coordinate-dependent token feature.
\item We verify the method on extensive simulated and clinical data. The studies support rapid lesion localization through tooth and pulp segmentation and show generalization to other objects.
\end{enumerate}

The remainder of this paper is organized as follows. Section II reviews related work. Section III specifies the pool and two reads. Sections IV--V report comparisons, ablations, clinical scans, and other objects. Sections VI--VII interpret and conclude.
\section{Related Works}
\subsection{Learning-based Reconstruction}
In this line of work, learning-based reconstruction replaces a hand-crafted penalty with a data-driven prior, so a volume can be obtained in one forward pass or a few unrolled steps. Image-domain networks such as FBPConvNet correct an FDK volume after it has been formed~\cite{jin2017deep}. Their cost is low, yet the input already contains streaks, and fine structure is easily smoothed away. Methods that use projections and images jointly keep a measurement-consistency constraint. DRONE exchanges information between the two domains~\cite{wu2021drone}, and LEARN and learned primal-dual reconstruction learn the successive solver updates~\cite{chen2018learn,adler2018primal}. These methods still unroll several steps, so runtime grows with the step count and the learned updates often depend on the training geometry. A sampling-diffusion model covers several sparsity levels with one network, but its sampling remains iterative~\cite{yang2025ctsdm}.

Cone-beam acquisition extends the problem from slices to three-dimensional coordinates. DIF-Net, C$^{2}$RV, and geometry-aware attenuation learning place projection information at the corresponding coordinates~\cite{lin2023difnet,lin2024c2rv,liu2025geometry}. Once global context is written into this representation, later coordinate queries rely mainly on local features. DeepSparse adds cross-scale pretraining and target-domain finetuning, which still requires large-scale data from many sources~\cite{lin2026deepsparse}. An off-the-shelf tooth and pulp segmenter can separate structures used for root-canal localization only when the input boundaries are already reliable~\cite{zhong2025tips}.

\begin{figure*}[!t]
\centering
\includegraphics[width=\textwidth]{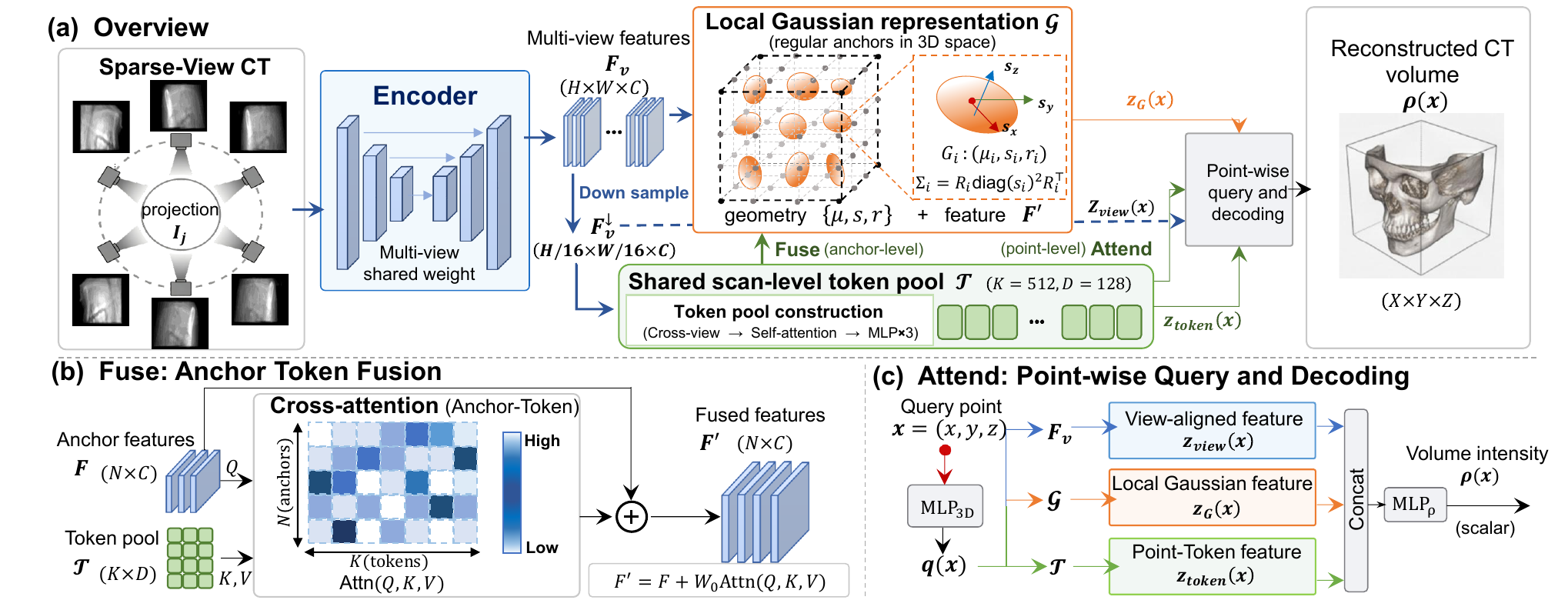}
\caption{Architecture of SparseTokenGS. Sparse-view projections enter a weight-shared encoder, which keeps a full-resolution map for view alignment and a downsampled map for the anchors and the token pool. The pool is then read twice. In (a), Fuse updates the anchors and Attend updates each query, and the decoded intensities form the volume. In (b), each anchor queries the token pool and a zero-initialized residual is added to its local feature. In (c), a coordinate network embeds the query, and the view feature, the neighboring Gaussian feature, and the point-token feature are concatenated and mapped to one intensity.}
\label{fig:method}
\end{figure*}

\subsection{Neural Fields, Gaussians, and Tokens}
The methods above still reconstruct on a discrete grid, or write global context only once during field construction. Neural fields and Gaussians represent the volume by a continuous function or a small set of primitives and do not store a dense voxel grid, which reduces computation and strengthens local structure. Neural radiance fields represent a scene with a coordinate network~\cite{mildenhall2021nerf}. NAF, SAX-NeRF, and NeAT fit an attenuation field to a single scan~\cite{zha2022naf,cai2024saxnerf,ruckert2022neat}, and DiffNR adds a diffusion prior~\cite{su2026diffnr}. Each patient still requires a separate optimization.

Radiative Gaussian splatting, R$^{2}$-Gaussian, and DGR likewise optimize these primitives for each scan~\cite{cai2024xgaussian,zha2024r2gaussian,wu2025dgr}.  

Among feed-forward methods, DIF-Gaussian predicts Gaussian features from a volumetric loss, with the projection loss used only as optional test-time refinement~\cite{lin2024difgaussian}, and X-GRM regresses a Gaussian at each voxel in one Transformer pass~\cite{liu2025xgrm}. In these designs the appearance of the scan is written into the anchors or voxels when the field is constructed, so later queries can only interpolate the local features.

An individual Gaussian describes only its neighborhood, so evidence from the acquired views must first be aggregated. Recent methods compress this evidence into tokens and then build the Gaussian field from them. After X-LRM lifts the tokens into a triplane, coordinate queries no longer access them~\cite{zhang2026xlrm}. Novel view synthesis in computer vision likewise uses learnable tokens to predict 3D Gaussians. TokenGS predicts Gaussian centers from tokens and does not keep them as a conditioning pool for later queries~\cite{ren2026tokengs}. Once the field is formed, neither the anchors nor the query coordinates can read the original tokens.

\section{Methodology}
\label{sec:method}
After a feed-forward Gaussian field is generated, the appearance of a scan is difficult to recover at a later coordinate query if it has been stored only as dispersed local features. SparseTokenGS therefore introduces a reusable scan-level token pool that preserves the global information until decoding. This section first describes the two reads of the pool, at the Gaussian anchors and at the query coordinates, and then explains the construction of the pool, the parameterization of the anchors, and the training of the hybrid query with volumetric supervision alone.

\subsection{Dual-Level Token Conditioning}
Given projections $\{I_v\}_{v=1}^{V}$ and their geometry, SparseTokenGS predicts a normalized field $\widehat{\rho}(\mathbf{x})$ for $\mathbf{x}\in[0,1]^3$. Its central operation reuses one scan-specific token matrix $T$ at two levels:
\begin{equation}
\begin{aligned}
F' &= F + W_0\,\mathcal{A}_{a}(F,T,T),\\
z_{\mathrm{token}}(\mathbf{x}) &=
\mathrm{LN}\!\left(q(\mathbf{x})+
\mathcal{A}_{p}(q(\mathbf{x}),T,T)\right).
\end{aligned}
\label{eq:dual}
\end{equation}
Here $F$ contains the local Gaussian-anchor features, $W_0$ is a zero-initialized linear map, $q$ embeds the query coordinate, and $\mathcal{A}_{a}$ and $\mathcal{A}_{p}$ are cross-attention blocks with separate parameters. In each block the first argument is the query, and the two copies of $T$ supply the keys and the values. Layer normalization (LN) is applied to the point-token update. The first read updates the anchor features that are later aggregated over spatial neighbors. The second read does not pass through that aggregation. It supplies a contextual feature that depends on the query coordinate. The two reads share the pool $T$ but not their attention parameters. Fig.~\ref{fig:method} shows the same flow. In (a), the weight-shared encoder keeps a full-resolution map for view alignment and a downsampled map for the anchors and the pool. Fuse reads the pool at the anchors, and Attend reads it at the query. In (b), each anchor is one row of the attention map and each token is one column. In (c), a coordinate multilayer perceptron (MLP) embeds $\mathbf{x}$, and $\mathrm{MLP}_{\mathrm{dec}}$ maps the view, Gaussian, and point-token features to one intensity. In the figure, $\mathcal{T}$ denotes $T$, $z_G$ denotes $z_{\mathrm{gs}}$, and $\mathrm{MLP}_{\rho}$ denotes $\mathrm{MLP}_{\mathrm{dec}}$.

Both reads use the same encoder. The U-Net~\cite{ronneberger2015unet}, with weights shared across views, produces two maps for each view. The full-resolution map $F_v\in\mathbb{R}^{128\times H\times W}$ is used for view alignment, and the bottleneck map $B_v\in\mathbb{R}^{1024\times H/16\times W/16}$ is used for the token pool and the anchors. Let $\pi_v(\mathbf{x})$ denote the detector coordinate of $\mathbf{x}$ under the known geometry. Bilinear sampling at $\pi_v$ aligns each map with a three-dimensional position. The prediction target is normalized image intensity, and linear attenuation is not calibrated.

\subsection{Scan-Specific Token Pool}
The scan-specific pool compresses the bottleneck features of all views into a fixed-length appearance representation. A $1\times1$ convolution reduces the bottleneck channels to $d=128$. For one scan, the tensor has channel-first layout $U\in\mathbb{R}^{V\times d\times H_b\times W_b}$, with $H_b=H/16$ and $W_b=W/16$. These features are reshaped in the same order into $M_{\mathrm{raw}}\in\mathbb{R}^{VH_bW_b\times d}$. Uniformly spaced row indices then retain at most $2048$ rows as the appearance memory $M$. Camera pose, view index, and two-dimensional positional embeddings are not added.

A learned matrix $T^{(0)}\in\mathbb{R}^{512\times d}$, initialized with Gaussian noise of standard deviation $0.02$, queries this memory. A linear map produces keys $K_M$ and values $V_M$. Three decoder blocks then refine the tokens. Each block applies cross-attention to $(K_M,V_M)$, followed by pre-normalized self-attention $\mathcal{S}_{\ell}$ and a pre-normalized feed-forward block $\mathcal{F}_{\ell}$ with expansion ratio $4$ and a Gaussian error linear unit (GELU). All three operations are residual updates. Cross-attention uses four heads of dimension $d_h=32$. Within each head, the projected query is layer-normalized before scaled dot-product attention. The final pool is $T=\mathrm{LN}(T^{(3)})$.

Each scan builds its own $T$ from the current projections. The initial tokens and the attention weights are shared across scans. The pool length does not depend on the number of views or output points, so its storage is fixed.

\begin{figure}[!t]
\centering
\includegraphics[width=\columnwidth]{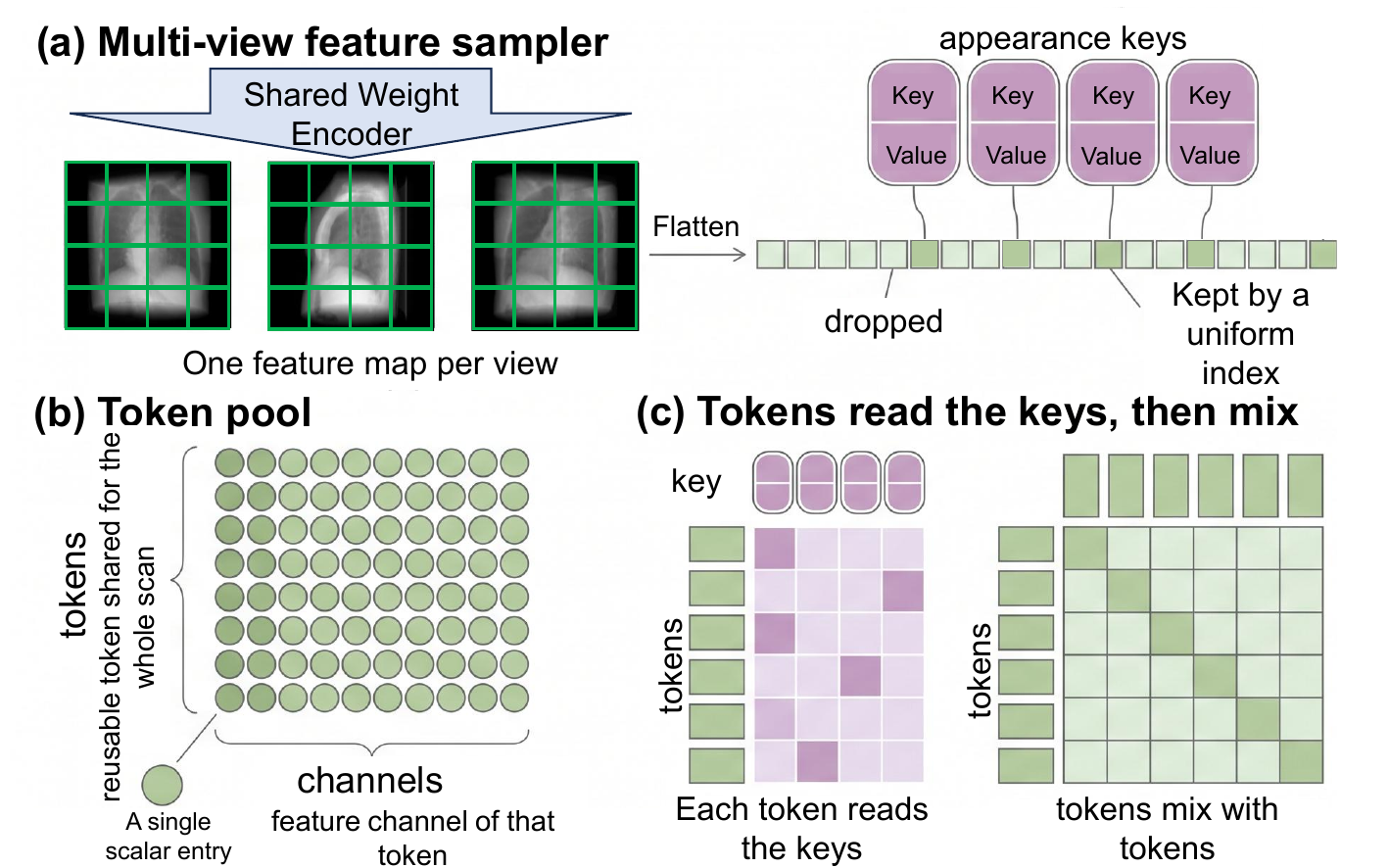}
\caption{Construction of the scan-specific token pool. (a) Multi-view feature sampling. (b) The token matrix. (c) Cross-attention and self-attention. The grids are schematic.}
\label{fig:pool}
\end{figure}

Fig.~\ref{fig:pool} shows the same construction. In (a), one feature map per view is flattened, and uniformly spaced cells are kept as keys and values. In (b), each row is one token and each entry is one channel. In (c), the tokens first read those keys and then mix with one another. A feed-forward update follows this mixing.

\subsection{Token-Conditioned Gaussian Anchors}
A lattice of resolution $R=14$ contains $A=R^3=2744$ nodes $\mathbf{c}_{ijk}=(i,j,k)/R$, where $i,j,k\in\{0,\ldots,R-1\}$. At node $\mathbf{c}_j$, local evidence is the channelwise maximum, over views, of the bottleneck samples $B_v$ at $\pi_v(\mathbf{c}_j)$. The maximum retains the strongest response among the views at that node. Two separate MLPs predict the anchor feature $f_j$ and the raw geometry. The ten geometry channels comprise a three-dimensional offset, a three-dimensional scale, and a quaternion.

The offset and scale logits are multiplied by $\tau=512$ before the sigmoid; the quaternion is normalized separately. With the rescaled outputs written as $\widehat{\delta}_j,\widehat{\mathbf{s}}_j,\widehat{r}_j$, the center and scale are
\begin{equation}
\begin{aligned}
\delta_j&=\big(\sigma(\widehat{\delta}_j)-\tfrac12\big)/R,\\
\mu_j&=\mathbf{c}_j+\delta_j,\\
 \mathbf{s}_j&=R^{-1/2}\big(0.7+0.6\,\sigma(\widehat{\mathbf{s}}_j)\big).
\end{aligned}
\label{eq:offset}
\end{equation}
Here $\sigma$ is the sigmoid function. The offset on each axis is at most half a lattice step, and each scale component lies between $0.7R^{-1/2}$ and $1.3R^{-1/2}$. The quaternion $\widehat{r}_j$ is normalized to unit length and defines the rotation $\mathcal{R}_j$. The covariance is $\Sigma_j=\mathcal{R}_j\,\mathrm{diag}(\mathbf{s}_j)^2\,\mathcal{R}_j^\top+10^{-6}I$. The offset bound limits only the center. It does not limit the Gaussian support, so neighboring kernels may overlap.

The features $f_j$ are stacked into $F\in\mathbb{R}^{A\times d}$. Anchor Token Fusion then applies the first line of \eqref{eq:dual}. The weight and bias of $W_0$ are both initialized at zero, so the fused anchor features initially equal the local prediction, while the point-token branch is already active. The geometry head receives no token, so the center, scale, and rotation stay tied to the local projection evidence, while appearance is the part conditioned on the pool. This head uses only those local features, although it shares the encoder with the token branch.

\subsection{Hybrid Point Query}
The decoder combines three $128$-dimensional features. The view-aligned branch forms $z_{\mathrm{view}}(\mathbf{x})$ as the channelwise maximum, over views, of the full-resolution samples $F_v$ at $\pi_v(\mathbf{x})$. For the Gaussian branch, $\mathcal{N}_5(\mathbf{x})$ contains the five nearest unshifted lattice nodes $\mathbf{c}_j$. The neighborhood is selected on the original lattice, and the density at these neighbors uses the shifted centers $\mu_j$:
\begin{equation}
w_j(\mathbf{x})=
\frac{\exp\!\left[-\tfrac12(\mathbf{x}-\mu_j)^\top
\Sigma_j^{-1}(\mathbf{x}-\mu_j)\right]}
{(2\pi)^{3/2}|\Sigma_j|^{1/2}}.
\label{eq:gauss}
\end{equation}
The selected weights are not normalized to sum to one, so the kernel magnitude remains in the aggregated feature. The Gaussian feature is
\begin{equation}
z_{\mathrm{gs}}(\mathbf{x})=
\mathrm{LeakyReLU}\!\left(
\sum_{j\in\mathcal{N}_5(\mathbf{x})}w_j(\mathbf{x})f'_j\right).
\label{eq:knn}
\end{equation}
Here $f'_j$ is row $j$ of $F'$. Tokens affect this branch only through the aggregated anchor features.

A two-layer coordinate MLP with GELU maps $\mathbf{x}$ to $q(\mathbf{x})\in\mathbb{R}^d$. Point-Token Attention then forms $z_{\mathrm{token}}(\mathbf{x})$ by the second line of \eqref{eq:dual}. Its query depends only on the coordinate, so this branch supplies a scan-level feature in addition to the two local branches. Every query coordinate reads the same pool $T$, while the attention weights may vary with the coordinate.

The concatenated vector has $384$ channels:
\begin{equation}
\widehat{\rho}(\mathbf{x})=
\mathrm{MLP}_{\mathrm{dec}}\!\left(
[z_{\mathrm{view}}(\mathbf{x}),z_{\mathrm{gs}}(\mathbf{x}),
z_{\mathrm{token}}(\mathbf{x})]\right).
\label{eq:query}
\end{equation}
The decoder uses pointwise convolutions of widths $256$, $64$, $16$, and $1$, with batch normalization and LeakyReLU. Skip connections carry the concatenated input into the later stages. At evaluation, the prediction is clipped to $[0,1]$.

The main simulated-data models minimize
\begin{equation}
\begin{aligned}
\mathcal{L}_{\mathrm{vol}}&=
0.85\,\mathcal{L}_{\mathrm{MSE}}+0.15\,\mathcal{L}_{1},\\
\mathcal{L}&=\mathcal{L}_{\mathrm{vol}}
+0.25\,\mathcal{L}_{\mathrm{sim}}
+10^{-5}\,\mathrm{mean}|T|.
\end{aligned}
\label{eq:loss}
\end{equation}
Here $\mathcal{L}_{\mathrm{MSE}}$ is the mean squared error (MSE) and $\mathcal{L}_1$ is the mean absolute error of the sampled intensities. Each training sample first selects a $64^3$ spatial block and then draws $19683=27^3$ points at random. The predictions and reference values are arranged in this sampling order as $27^3$ arrays for $\mathcal{L}_{\mathrm{sim}}$. The term equals one minus a windowed similarity score. The score uses $7^3$ average pooling and the constants $(0.01)^2$ and $(0.03)^2$, following the form of the structural similarity index (SSIM)~\cite{wang2004ssim}. $\mathcal{L}_{\mathrm{sim}}$ is therefore distinct from the spatial SSIM computed on the reconstructed volume at evaluation. Training includes no projection-consistency term.

\section{Experimental Setting}
\label{sec:experiments}
\subsection{Simulated Experiments and Datasets}
We conducted quantitative evaluation on the chest computed tomography (CT) dataset LUNA16~\cite{setio2017luna16} and the dental CBCT datasets ToothFairy~\cite{bolelli2025toothfairy} and ToothFairy4~\cite{toothfairy4}. LUNA25~\cite{luna25challenge} is shown with these cohorts in Fig.~\ref{fig:four}. Each quantitative dataset is split into training, validation, and test scans at a ratio of about $8{:}1{:}1$. Before projection, chest intensities are clipped to $[-1000,1000]$ Hounsfield units (HU) and dental intensities to $[-1000,3000]$ HU. The volumes are then normalized to $[0,1]$ and resampled to $256^3$. Cone-beam projections are simulated from these normalized volumes for chest and dental anatomy. Display windows follow the anatomy: chest slices use $[-1000,1000]$ HU and dental slices use $[-800,1200]$ HU.

For each quantitative dataset, we generate $50$ projections along a fixed $180^\circ$ circular arc that starts at $0^\circ$. The chest geometry uses source-to-isocenter and source-to-detector distances of $800$ mm and $1200$ mm, a $256^2$ detector, and $2.4$ mm pixels. The dental geometry uses source-to-isocenter and source-to-detector distances of $900$ mm and $1000$ mm, with detector spacings of about $0.209$ mm and $0.543$ mm. Every angular subset is taken from this same arc. We trained SparseTokenGS models at 3, 6, 10, 14, and 18 views. During training, projection indices are drawn at approximately uniform spacing with a random starting offset. At test time, the indices are fixed and the offset is zero.

\subsection{Baselines and Evaluation Protocol}
In the quantitative comparison, we include FDK, SART, FBPConvNet, DIF-Net, feed-forward DIF-Gaussian without test-time refinement, and X-GRM~\cite{feldkamp1984practical,andersen1984sart,jin2017deep,lin2023difnet,lin2024difgaussian,liu2025xgrm}. Fig.~\ref{fig:four} additionally shows the per-scan methods DGR, DiffNR, and R$^{2}$-Gaussian~\cite{wu2025dgr,su2026diffnr,zha2024r2gaussian}. The learned feed-forward baselines are trained separately for each dataset and each view count. SART runs for $20$ iterations with the simulator projector. FBPConvNet is a residual U-Net trained for eight epochs on axial FDK slices from $80$ scans. We compute the peak signal-to-noise ratio (PSNR), SSIM, and root-mean-square error (RMSE) with a data range of $1$. The evaluation SSIM uses a $7^3$ uniform window on the reconstructed volume, which is separate from the sample-order term used in training.

\subsection{Clinical Data and Other Objects}
We also evaluate the method on clinical dental CBCT and on volcanic rock and chicken. The retrospective dental scans use a LargeV SMART3D system at $100$ kV and $6$ mA. The nominal source-to-isocenter and source-to-detector distances are $420$ mm and $640$ mm. The detector has $1212\times1392$ pixels at $0.1$ mm spacing, and the scanner metadata lists $676$ projections per rotation. The archive stores projection indices. The full-view reconstructions from the scanner, originally $488\times488\times332$ voxels at $0.25$ mm, are resampled to $256^3$. The resulting spacing is about $0.477$ mm in-plane and $0.324$ mm along the longitudinal axis, and we have obtained the relevant ethical approval for these retrospective scans.

Volcanic rock and chicken are included to examine generalization to objects outside the chest and dental anatomy. Both acquisitions use a source-to-detector distance of $800$ mm and a detector element size of $0.139$ mm. The volcanic rock scan has $1440$ projections and $2531$ detector columns, and the chicken scan has $1441$ projections and $2568$ detector columns. The reconstruction reports an isocenter voxel size of $0.062$ mm. The exported ROI reference volumes use isotropic sampling of $15.6375$ $\mu$m for the rock and $19.1125$ $\mu$m for the chicken. This experiment uses eighteen views. Adaptation uses augmented copies and reference volumes of the same two specimens.

\subsection{Implementation and Computational Cost}
We implement SparseTokenGS and the learned comparison methods in PyTorch. We perform training, validation, and testing on an Intel Xeon Platinum 8558 (4.00 GHz) central processing unit (CPU) and eight NVIDIA H100 graphics processing units (GPUs). In the simulated experiments, SparseTokenGS and the feed-forward baselines DIF-Net, DIF-Gaussian, and X-GRM are each trained for $600$ epochs, with one checkpoint for every dataset and every view count. SparseTokenGS uses a batch size of $2$ and stochastic gradient descent (SGD) with momentum $0.98$. Its learning rate starts at $0.01$ and decays to $5\times10^{-4}$ by the last epoch, with weight decay $3\times10^{-4}$. FBPConvNet is trained for eight epochs. The per-scan methods are optimized case by case. In the six-view experiments, DGR, DiffNR, and R$^{2}$-Gaussian use $300$, $12000$, and $30000$ steps, respectively. The network uses $512$ tokens, three attention blocks, four heads, at most $2048$ memory rows, a $14^3$ anchor lattice, and five neighbors at each query. The loss in \eqref{eq:loss} is computed on $19683$ sampled points. Validation is performed on a half-resolution grid, and the final volumes are evaluated at $256^3$.

\begin{figure}[!t]
\centering
\includegraphics[width=\columnwidth]{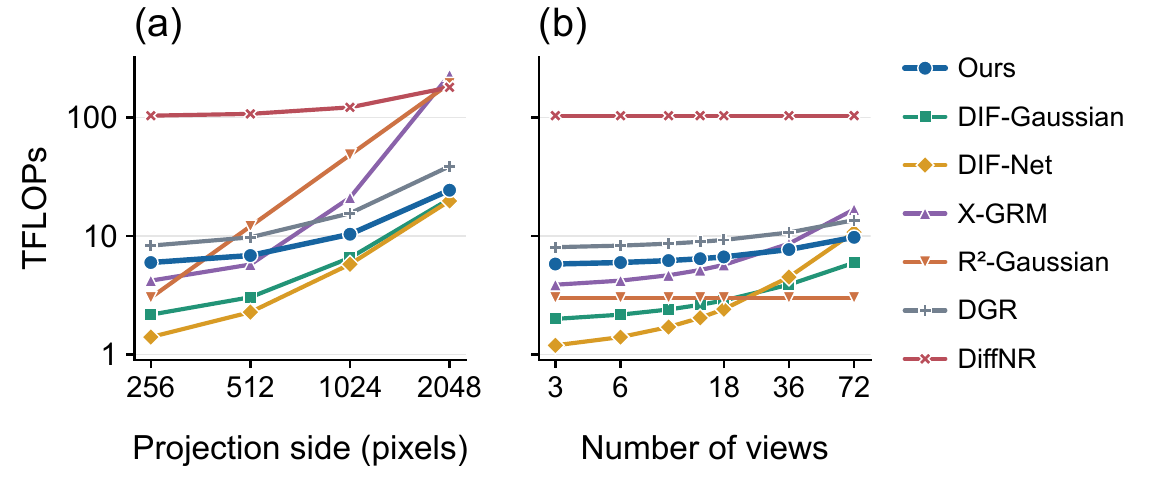}
\caption{Analytical MAC estimates for a $256^3$ volume, using the same $10^{12}$-MAC unit as Table~\ref{tab:cost}. (a) Six views as the projection side grows. (b) Cost as the number of views changes for the per-scan methods in that estimate.}
\label{fig:flops}
\end{figure}

Table~\ref{tab:cost} gives analytical multiply-accumulate (MAC) estimates for six $256^2$ projections and a $256^3$ output, including the learned comparison methods and the per-scan methods. One MAC counts as one multiplication together with its accumulation. The estimates leave out batch normalization and treat interpolation as four taps per channel. They describe arithmetic cost. MAC counts for the per-scan methods include the iteration budget of the six-view experiments, as in Fig.~\ref{fig:flops}. Fig.~\ref{fig:flops} uses the same $10^{12}$-MAC unit for other detector sizes, other view counts, and three per-scan methods.

\begin{table}[!t]
\caption{Parameters and MAC estimates for six $256^2$ projections and a $256^3$ volume. T denotes $10^{12}$ MACs. DGR, DiffNR, and R$^{2}$-Gaussian include the six-view iteration budget.}
\label{tab:cost}
\centering
\scriptsize
\begin{tabular}{lrr}
\toprule
Method & Parameters (M) & MACs (T)\\
\midrule
FBPConvNet & 1.9 & 2.4\\
DGR & 7.1 & 8.3\\
DiffNR & 1305.5 & 103.6\\
R$^{2}$-Gaussian & 0.97 & 3.0\\
X-GRM & 172.6 & 4.2\\
DIF-Net & 31.1 & 1.4\\
DIF-Gaussian & 31.7 & 2.2\\
SparseTokenGS & 32.8 & 6.0\\
\bottomrule
\end{tabular}
\end{table}

The cost of token-to-memory attention grows as $O(LKMd)$ over $L=3$ layers, with $K=512$ and $M\leq2048$. Anchor-token attention grows as $O(AKd)$, and point-token attention grows as $O(QKd)$ for $Q$ output coordinates. Because the pool has a fixed length, both its storage and the number of tokens read by each query stay constant. The query branch still becomes more expensive as the output resolution increases, which accounts for its higher arithmetic cost relative to DIF-Gaussian.

\section{Results}
\label{sec:results}
\subsection{Main Results}
\begin{table*}[t]
\caption{PSNR (dB) / SSIM / RMSE on the simulated chest and dental data. Bold marks the best in each row. Ours denotes SparseTokenGS.}
\label{tab:views}
\centering
\scriptsize
\setlength{\tabcolsep}{2.2pt}
\begin{tabular}{ccccccccc}
\toprule
\textbf{Dataset} & \textbf{View} & \textbf{FDK} & \textbf{SART} & \textbf{FBPConvNet} & \textbf{X-GRM} & \textbf{DIF-Net} & \textbf{DIF-Gaussian} & \textbf{Ours} \\
\midrule
\textbf{LUNA16} & 3 & 14.40/0.2644/0.1905 & 18.81/0.5549/0.1157 & 19.44/0.5872/0.1076 & 19.24/0.6926/0.1091 & 22.41/0.7082/0.0758 & 26.06/0.8254/0.0498 & \textbf{27.52}/\textbf{0.8602}/\textbf{0.0421} \\
 & 6 & 15.17/0.2380/0.1744 & 23.04/0.6741/0.0711 & 22.26/0.6683/0.0777 & 24.07/0.7775/0.0626 & 23.98/0.7178/0.0633 & 28.16/0.8573/0.0391 & \textbf{30.11}/\textbf{0.8933}/\textbf{0.0319} \\
 & 10 & 16.30/0.2483/0.1531 & 25.10/0.7393/0.0561 & 23.86/0.6345/0.0648 & 24.86/0.7967/0.0571 & 26.42/0.7587/0.0477 & 29.77/0.8809/0.0325 & \textbf{31.70}/\textbf{0.9103}/\textbf{0.0264} \\
 & 14 & 16.80/0.2621/0.1445 & 26.42/0.7781/0.0483 & 24.45/0.7134/0.0612 & 23.28/0.7843/0.0686 & 27.08/0.7763/0.0443 & 30.33/0.8872/0.0305 & \textbf{32.32}/\textbf{0.9144}/\textbf{0.0242} \\
 & 18 & 17.14/0.2730/0.1389 & 27.33/0.8037/0.0434 & 25.06/0.7662/0.0569 & 24.55/0.7930/0.0592 & 27.91/0.7946/0.0402 & 30.91/0.8957/0.0285 & \textbf{32.99}/\textbf{0.9206}/\textbf{0.0224} \\
\midrule
\textbf{ToothFairy} & 3 & 14.13/0.1829/0.1965 & 22.80/0.6680/0.0735 & 22.42/0.5892/0.0768 & 20.17/0.6453/0.0980 & 22.12/0.6506/0.0783 & 25.81/0.7883/0.0512 & \textbf{27.55}/\textbf{0.8538}/\textbf{0.0419} \\
 & 6 & 15.68/0.2141/0.1644 & 24.98/0.7314/0.0572 & 24.09/0.6877/0.0637 & 24.51/0.7697/0.0595 & 25.22/0.7335/0.0548 & 28.39/0.8476/0.0381 & \textbf{30.62}/\textbf{0.9043}/\textbf{0.0299} \\
 & 10 & 16.89/0.2306/0.1430 & 26.71/0.7678/0.0469 & 26.09/0.7228/0.0500 & 22.98/0.7094/0.0709 & 27.15/0.7738/0.0439 & 30.36/0.8870/0.0303 & \textbf{32.57}/\textbf{0.9267}/\textbf{0.0235} \\
 & 14 & 17.84/0.2488/0.1283 & 27.92/0.7948/0.0408 & 27.28/0.8000/0.0444 & 23.47/0.7549/0.0671 & 27.83/0.7919/0.0406 & 31.05/0.8934/0.0280 & \textbf{33.39}/\textbf{0.9341}/\textbf{0.0214} \\
 & 18 & 18.53/0.2740/0.1185 & 29.03/0.8187/0.0359 & 28.08/0.6871/0.0400 & 23.65/0.7574/0.0657 & 28.69/0.8095/0.0368 & 31.46/0.8976/0.0267 & \textbf{34.18}/\textbf{0.9402}/\textbf{0.0195} \\
\midrule
\textbf{ToothFairy4} & 3 & 14.73/0.2445/0.1834 & 21.00/0.6102/0.0919 & 19.15/0.5290/0.1116 & 20.48/0.6348/0.0946 & 23.28/0.7084/0.0686 & 26.16/0.8176/0.0492 & \textbf{27.42}/\textbf{0.8583}/\textbf{0.0425} \\
 & 6 & 15.53/0.2059/0.1673 & 24.70/0.7303/0.0604 & 22.41/0.6462/0.0771 & 23.96/0.7599/0.0634 & 25.86/0.7584/0.0509 & 28.63/0.8680/0.0370 & \textbf{30.12}/\textbf{0.9010}/\textbf{0.0334} \\
 & 10 & 16.66/0.2195/0.1468 & 26.41/0.7682/0.0498 & 23.90/0.5778/0.0653 & 23.43/0.3969/0.0674 & 27.34/0.7815/0.0429 & 30.05/0.8880/0.0315 & \textbf{31.64}/\textbf{0.9150}/\textbf{0.0262} \\
 & 14 & 17.66/0.2385/0.1309 & 27.64/0.7960/0.0432 & 24.11/0.6605/0.0643 & 23.39/0.7461/0.0677 & 28.19/0.8040/0.0390 & 30.90/0.9002/0.0285 & \textbf{32.48}/\textbf{0.9236}/\textbf{0.0238} \\
 & 18 & 18.41/0.2614/0.1201 & 28.72/0.8186/0.0382 & 24.00/0.6946/0.0656 & 23.58/0.7607/0.0662 & 29.08/0.8169/0.0352 & 31.57/0.9065/0.0264 & \textbf{33.21}/\textbf{0.9290}/\textbf{0.0218} \\
\bottomrule
\end{tabular}
\end{table*}

As Table~\ref{tab:views} shows, six-view FDK PSNR stays around $15$ dB on the three datasets, since analytical backprojection from sparse angles leaves streaks in the volume. SART reduces the data residual by iteration and reaches about $23$ to $25$ dB, yet each case is optimized alone and low-contrast texture stays limited. FBPConvNet corrects an FDK volume that already contains streaks, reaching about $22$ to $24$ dB, and the boundaries it can recover are limited by that input. These three groups are respectively bounded by analytical backprojection, per-case iteration, and an already-streaked input, so none keeps both detail and within-scan consistency under six projections.

In the same table, X-GRM and DIF-Net write multi-view evidence into voxels or local coordinate features while the field is constructed, so a later query cannot retrieve the appearance of the whole scan. Their six-view PSNR stays around $24$ to $26$ dB. DIF-Gaussian keeps the local Gaussian prior and reaches about $28$ to $29$ dB, but the scan-level context is fixed once the anchors are formed. SparseTokenGS reads the same token pool at the query coordinates, reaching $30.11$ dB on LUNA16, $30.62$ dB on ToothFairy, and $30.12$ dB on ToothFairy4, which is $1.95$ to $2.23$ dB above DIF-Gaussian. The shared weakness of the feed-forward baselines is that scan-level appearance is lost once the field is constructed. SparseTokenGS keeps it until decoding, so each query can still recover the appearance of the whole scan, and the volume stays more consistent under sparse views.

\begin{figure}[!t]
\centering
\includegraphics[width=\columnwidth]{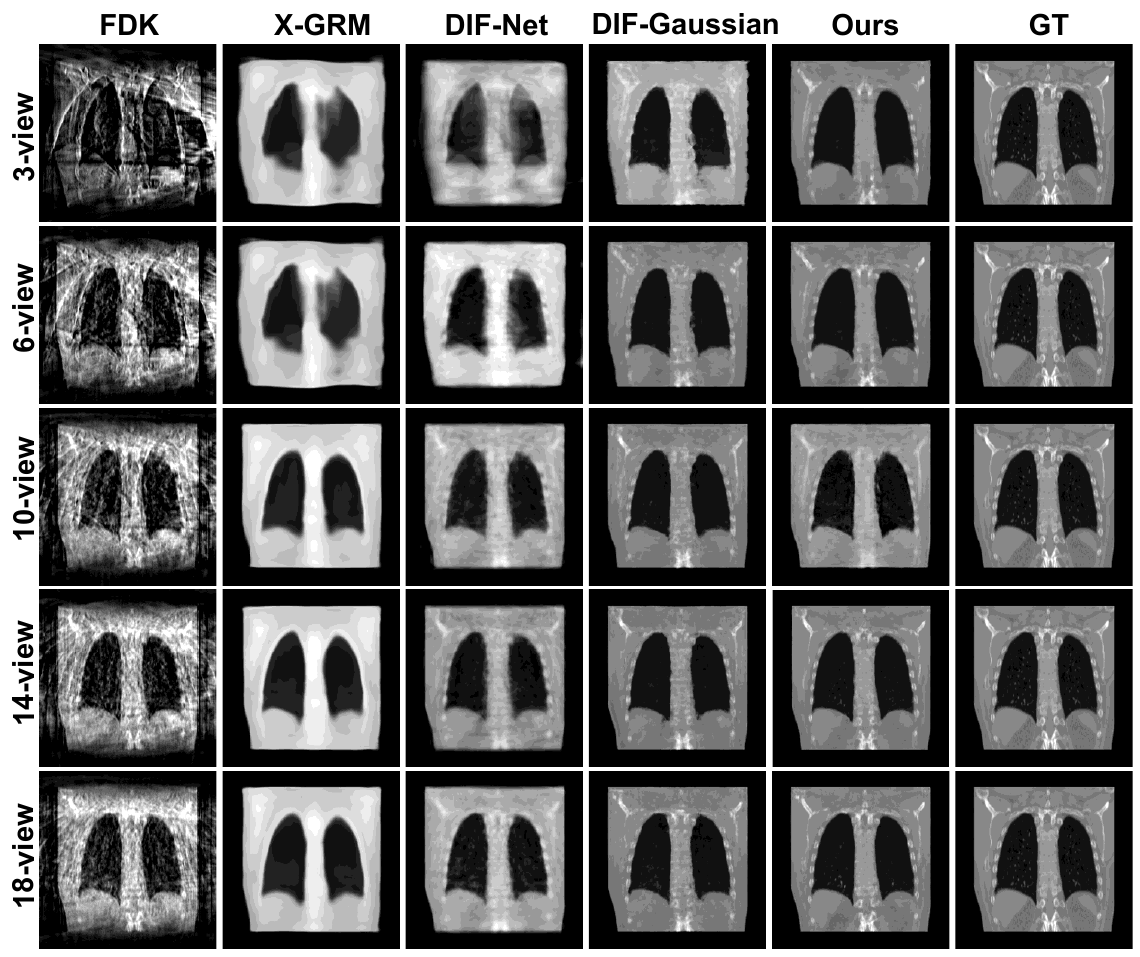}
\caption{Coronal slices of one LUNA16 scan at $3$, $6$, $10$, $14$, and $18$ views. Columns are FDK, X-GRM, DIF-Net, DIF-Gaussian, SparseTokenGS, and the reference. Display window: $[-1000,1000]$ HU.}
\label{fig:luna_cor}
\end{figure}

\begin{figure}[!t]
\centering
\includegraphics[width=\columnwidth]{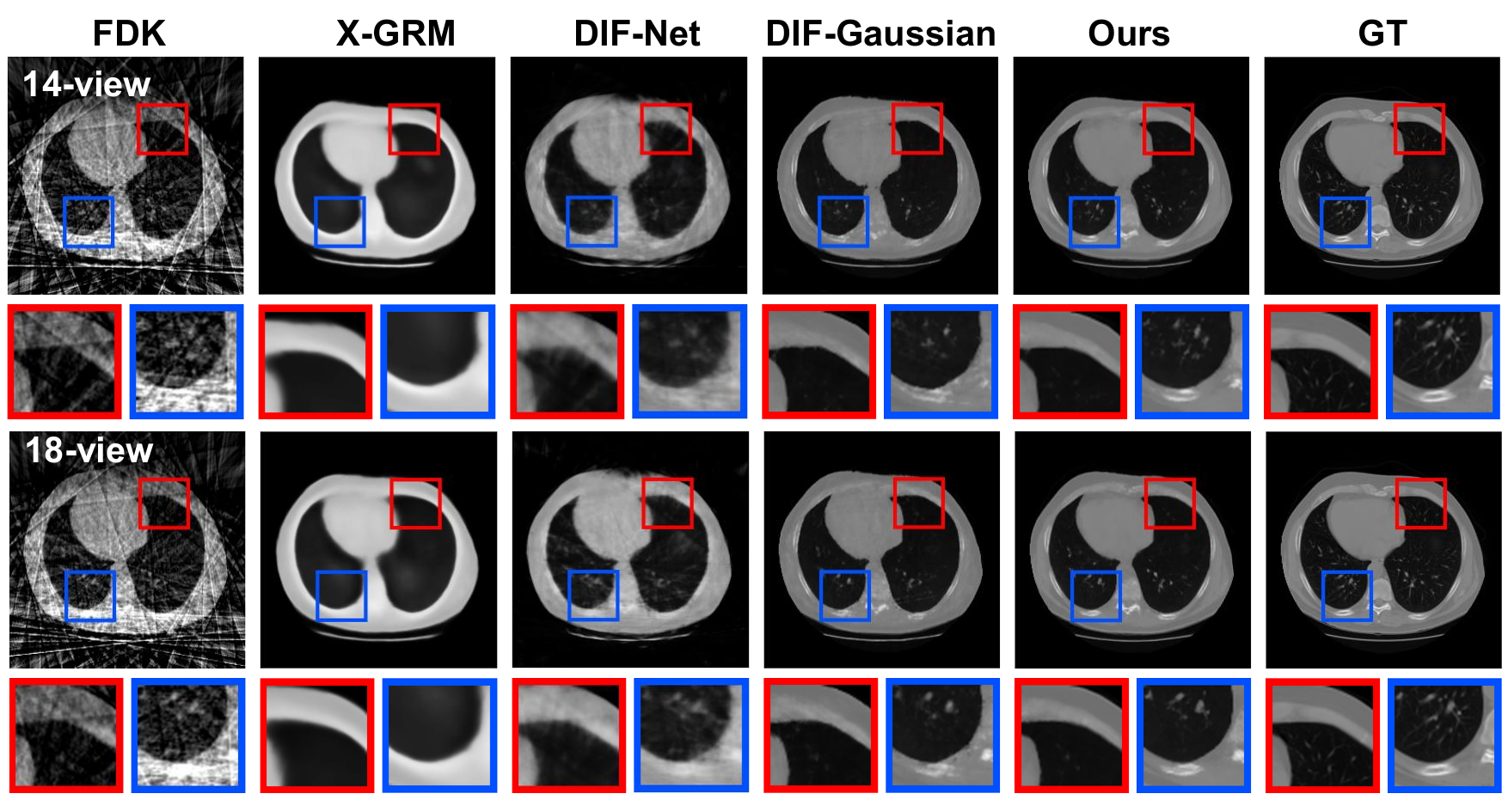}
\caption{Axial slices of the LUNA16 scan in Fig.~\ref{fig:luna_cor} at fourteen and eighteen views. Columns follow that figure. Colored boxes mark enlarged lung regions. Display window: $[-1000,1000]$ HU.}
\label{fig:luna_ax}
\end{figure}

As Figs.~\ref{fig:luna_cor} and \ref{fig:luna_ax} show, on this LUNA16 scan FDK still shows streaks, X-GRM and DIF-Net recover the main outline, DIF-Gaussian makes the boundaries clearer, and the chest wall and lung regions of SparseTokenGS move closer to the reference as the number of views grows. Methods that can still retrieve scan-level context from the pool are closer to the reference on fine boundaries, since a boundary seen in only a few views is otherwise lost in the neighbor average.

As Table~\ref{tab:views} shows, PSNR on LUNA16 rises from $27.52$ dB at three views to $32.99$ dB at eighteen views, while ToothFairy rises from $27.55$ to $34.18$ dB and ToothFairy4 from $27.42$ to $33.21$ dB. Because the same fixed-length pool receives fuller measurements as views increase, the advantage already present at three views remains, and at eighteen views SparseTokenGS still leads DIF-Gaussian by $2.08$ to $2.72$ dB. From three to eighteen views, RMSE falls from about $0.042$ to about $0.022$ on the three datasets. The number of views changes only the evidence written into the tokens, while the decoder and the field keep their sizes, so the same model structure serves three to eighteen views.

\begin{figure}[!t]
\centering
\includegraphics[width=\columnwidth]{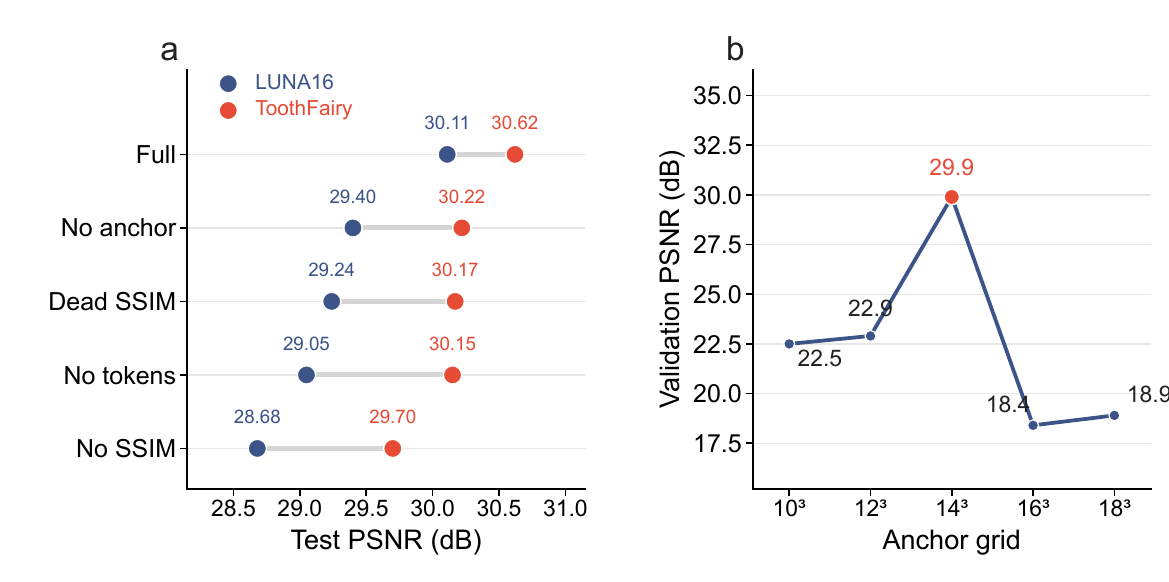}
\caption{Six-view ablation. (a) PSNR and SSIM on LUNA16 and ToothFairy for the settings in Table~\ref{tab:ablation}. (b) LUNA16 validation PSNR at epoch $400$ on the half-resolution grid, plotted against anchor-grid resolution.}
\label{fig:ablation}
\end{figure}

\begin{figure*}[t]
\centering
\includegraphics[width=0.92\textwidth]{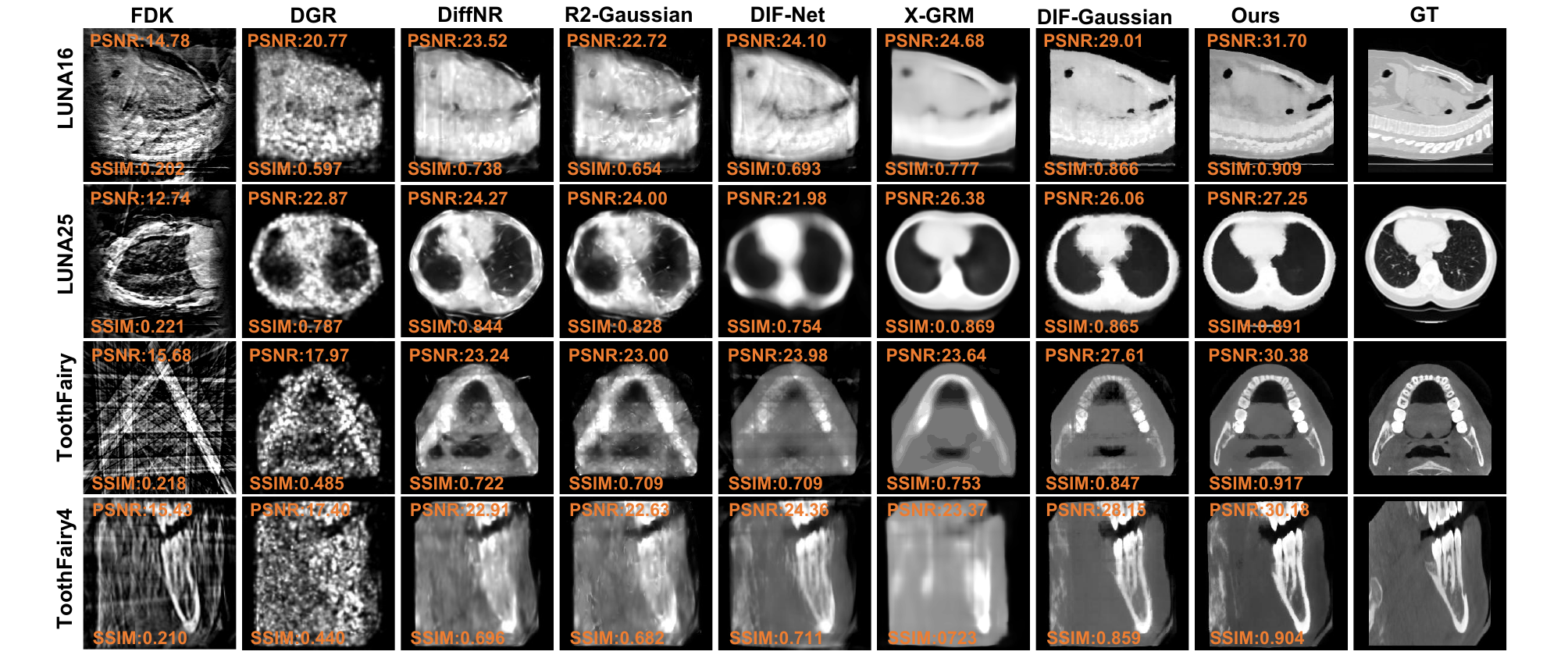}
\caption{Six-view reconstructions. Rows are LUNA16, LUNA25, ToothFairy, and ToothFairy4. Columns are FDK, DGR, DiffNR, R$^{2}$-Gaussian, DIF-Net, X-GRM, DIF-Gaussian, SparseTokenGS, and the reference. Numbers under the panels are PSNR (dB) and SSIM. Rows 1 and 2 use the chest window $[-1000,1000]$ HU. Rows 3 and 4 use the dental window $[-800,1200]$ HU.}
\label{fig:four}
\end{figure*}

As Fig.~\ref{fig:four} shows, each of the four cohorts contributes one scan, and the panel scores belong to that scan. DGR, DiffNR, and R$^{2}$-Gaussian optimize the scan separately, fitting appearance to that one case. DIF-Net, X-GRM, DIF-Gaussian, and SparseTokenGS use one forward pass, and SparseTokenGS retains the broad lung and dental contours because the query still reads the pool. The cohort means for LUNA16, ToothFairy, and ToothFairy4 are in Table~\ref{tab:views}.

\subsection{Ablation Results}
We ablate the two token reads, the similarity term, and the sample count on six-view LUNA16 and ToothFairy. As Table~\ref{tab:ablation} and Fig.~\ref{fig:ablation}(a) show, the main reference scores $30.1061$ dB and $0.8933$ on LUNA16, and $30.6234$ dB and $0.9043$ on ToothFairy. Removing anchor fusion lowers PSNR by $0.71$ dB and $0.41$ dB, since the anchors then miss one scan-level update. Removing the token pool at $20000$ samples lowers PSNR by $0.19$ dB and $0.01$ dB. Removing the similarity term lowers PSNR by $1.42$ dB and $0.92$ dB, because neighboring samples then lose their structural link. When only the anchor read is kept, the scores fall to $23.7512$ dB and $0.8016$, and $24.7824$ dB and $0.8096$, which is $6.35$ dB and $5.84$ dB below the main reference and the largest drop among the rows, because the five-neighbor average cannot replace the point-level read. Together, the point-level read accounts for the largest loss, and the similarity term and anchor fusion close smaller gaps.

\begin{table}[!t]
\caption{Six-view ablation on LUNA16 and ToothFairy. Each cell is PSNR (dB) / SSIM.}
\label{tab:ablation}
\centering
\scriptsize
\setlength{\tabcolsep}{3pt}
\begin{tabular}{lcc}
\toprule
Setting & LUNA16 & ToothFairy\\
\midrule
Main reference & 30.1061 / 0.8933 & 30.6234 / 0.9043\\
No anchor fusion & 29.3950 / 0.8877 & 30.2162 / 0.9028\\
20000 samples & 29.2359 / 0.8855 & 30.1663 / 0.9022\\
No tokens, 20000 & 29.0458 / 0.8825 & 30.1534 / 0.9019\\
No similarity term & 28.6847 / 0.8787 & 29.7013 / 0.8968\\
Anchor read only & 23.7512 / 0.8016 & 24.7824 / 0.8096\\
\bottomrule
\end{tabular}
\end{table}

As Fig.~\ref{fig:ablation}(b) shows, epoch-$400$ validation PSNR on the half-resolution grid is highest for the $14^3$ anchors. Denser grids cover the volume more finely, but the same training must fit more anchors, so their training loss is higher, and $14^3$ is the operating point between coverage and learnability.

\subsection{Clinical Dental Evaluation}
On the held-out clinical cases, SparseTokenGS reaches $29.51\pm0.67$ dB PSNR and $0.842\pm0.0095$ SSIM against the resampled full-view reconstruction, with sample standard deviations across cases. With six projections, the query coordinates can still read the scan-level context, so the dental arch and the mandible remain consistent as a whole. The pooled RMSE over all stored $256^3$ volumes, after division by $255$, is $0.0336$.

To characterize intensity-dependent error without assigning tissue labels, reference values of at least $0.325$ define a high-intensity region and values in $[0.05,0.325)$ an intermediate region. The corresponding mean absolute errors (MAEs) are $0.0556$ and $0.0208$. The high-intensity error is about $2.7$ times larger on average and larger in every case.

\begin{figure}[!t]
\centering
\includegraphics[width=\columnwidth]{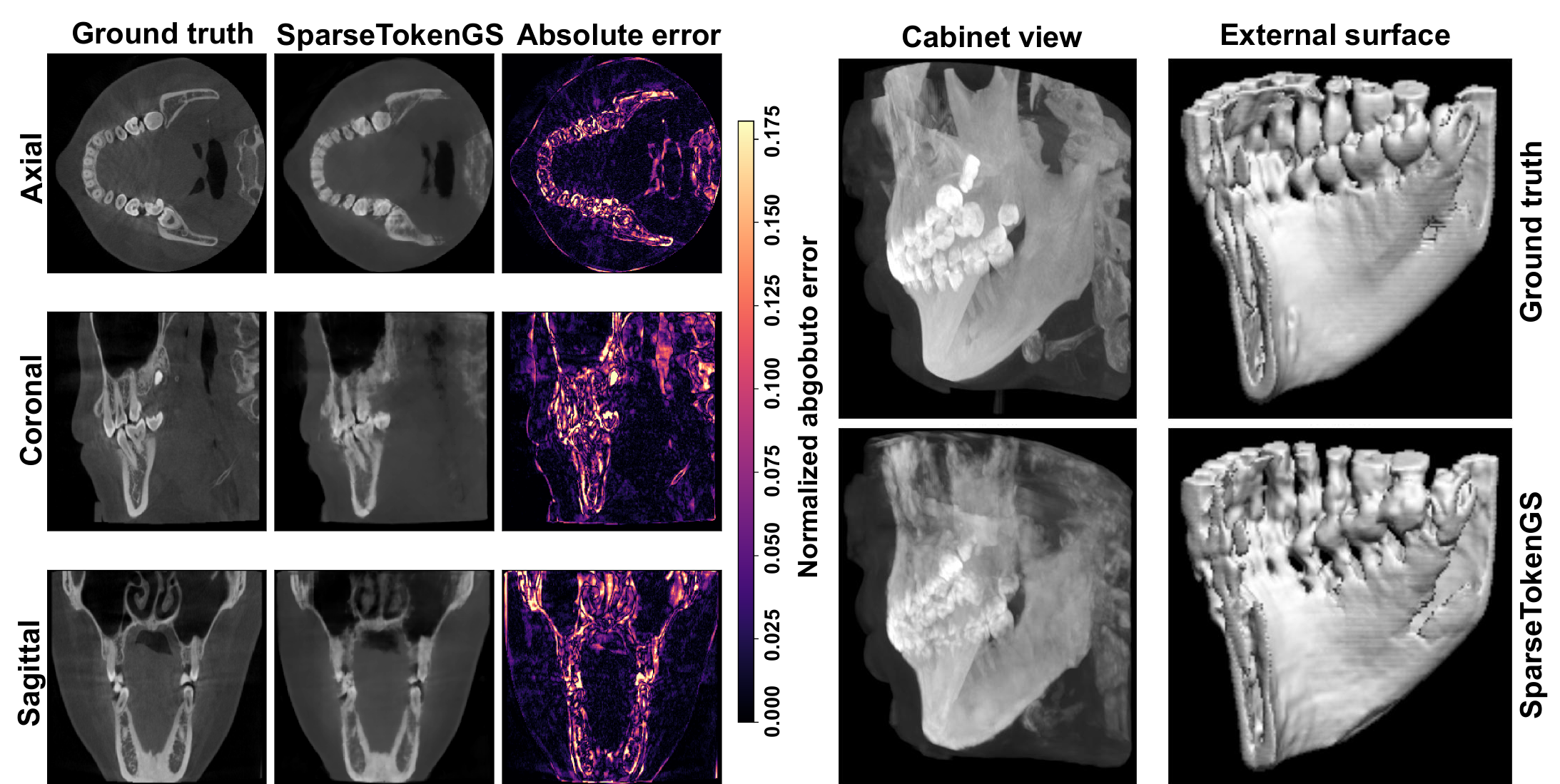}
\caption{Six-view reconstruction of one held-out clinical dental scan. Left: axial, coronal, and sagittal sections of the scanner reconstruction, SparseTokenGS, and absolute error. Right: reference and predicted external views.}
\label{fig:real}
\end{figure}

As Fig.~\ref{fig:real} shows, the dental arch and the mandibular outline agree with the full-view reference. Neighborhood aggregation at the anchors joins these outlines into a continuous contour, while the larger absolute error lies in the high-intensity region and follows the tooth and cortical boundaries, which are more sensitive to missing angles. Grayscale sections use the display window $[-800,1200]$ HU, and the printed scores are computed on stored volumes divided by $255$.

\begin{figure}[!t]
\centering
\includegraphics[width=\columnwidth]{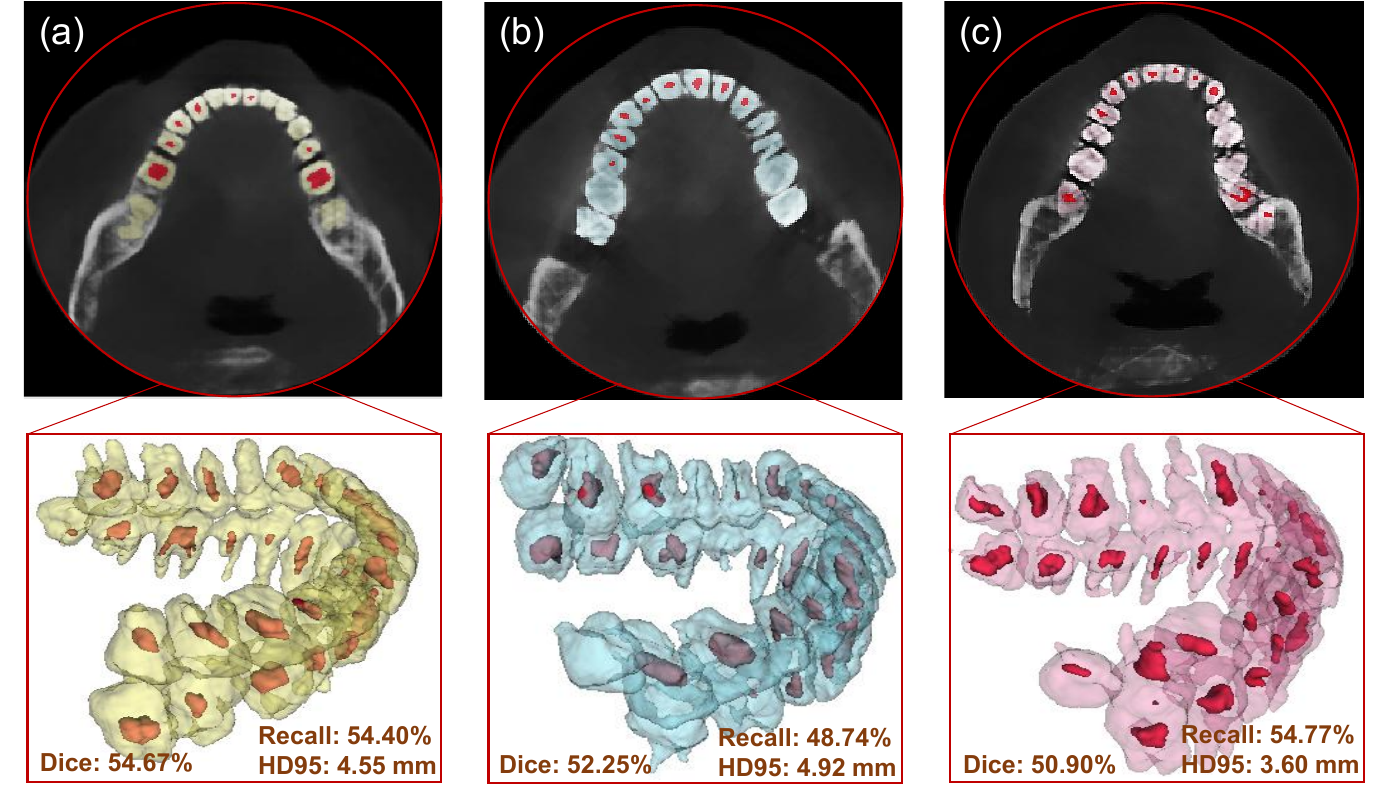}
\caption{TIPs on three SparseTokenGS dental reconstructions, without retraining~\cite{zhong2025tips}.}
\label{fig:tips}
\end{figure}

As Fig.~\ref{fig:tips} shows, the published TIPs model~\cite{zhong2025tips} is applied to three reconstructions without retraining. Across panels (a) to (c), the pulp Dice similarity coefficient (Dice) is 54.67\%, 52.25\%, and 50.90\%, and the 95th-percentile Hausdorff distance (HD95) is 4.55 mm, 4.92 mm, and 3.60 mm. Dice stays above $50\%$ and HD95 stays below $5$ mm on all three cases. In each panel the axial section lies above the three-dimensional rendering. The lighter color is the predicted tooth and the darker color is the predicted pulp. This delineation supports root-canal localization, and six projections are far fewer than the 676 acquired in a full rotation.

\subsection{Generalization to Different Objects}
\begin{figure}[!t]
\centering
\includegraphics[width=\columnwidth]{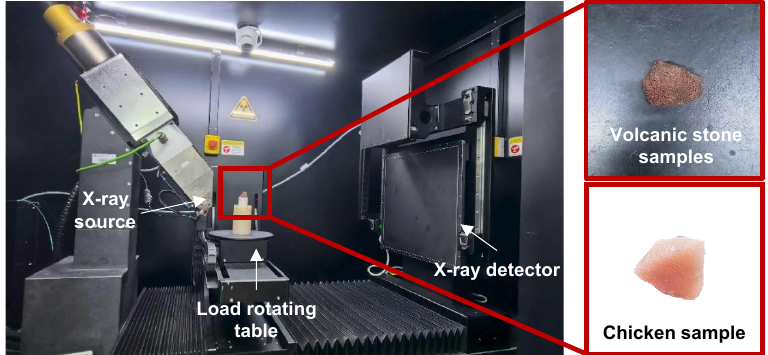}
\caption{Acquisition setup for the volcanic rock and chicken specimens. The eighteen-view reconstructions from this setup are reported in Table~\ref{tab:phantom} and Fig.~\ref{fig:phantom}.}
\label{fig:scanner}
\end{figure}

Volcanic rock and chicken lie outside the chest and dental anatomy used above, and Fig.~\ref{fig:scanner} shows the acquisition setup for these specimens. The four learned methods were adapted with the same supervised protocol on augmented copies and reference volumes of the two specimens.

\begin{table}[!t]
\caption{Eighteen-view PSNR (dB) / SSIM on volcanic rock and chicken. Ours denotes SparseTokenGS. Bold marks the highest value in each row.}
\label{tab:phantom}
\centering
\scriptsize
\setlength{\tabcolsep}{1.5pt}
\resizebox{\columnwidth}{!}{%
\begin{tabular}{lccccc}
\toprule
\textbf{Specimen} & \textbf{FDK} & \textbf{X-GRM} & \textbf{DIF-Net} & \textbf{DIF-Gaussian} & \textbf{Ours} \\
\midrule
Volcanic rock & 15.81/0.2724 & 19.79/0.5316 & 19.96/0.4874 & 20.24/0.3732 & \textbf{25.71}/\textbf{0.6802} \\
Chicken & 16.09/0.3109 & 21.21/0.5799 & 21.18/0.5630 & 21.50/0.4602 & \textbf{28.73}/\textbf{0.6861} \\
\bottomrule
\end{tabular}}
\end{table}

\begin{figure}[!t]
\centering
\includegraphics[width=\columnwidth]{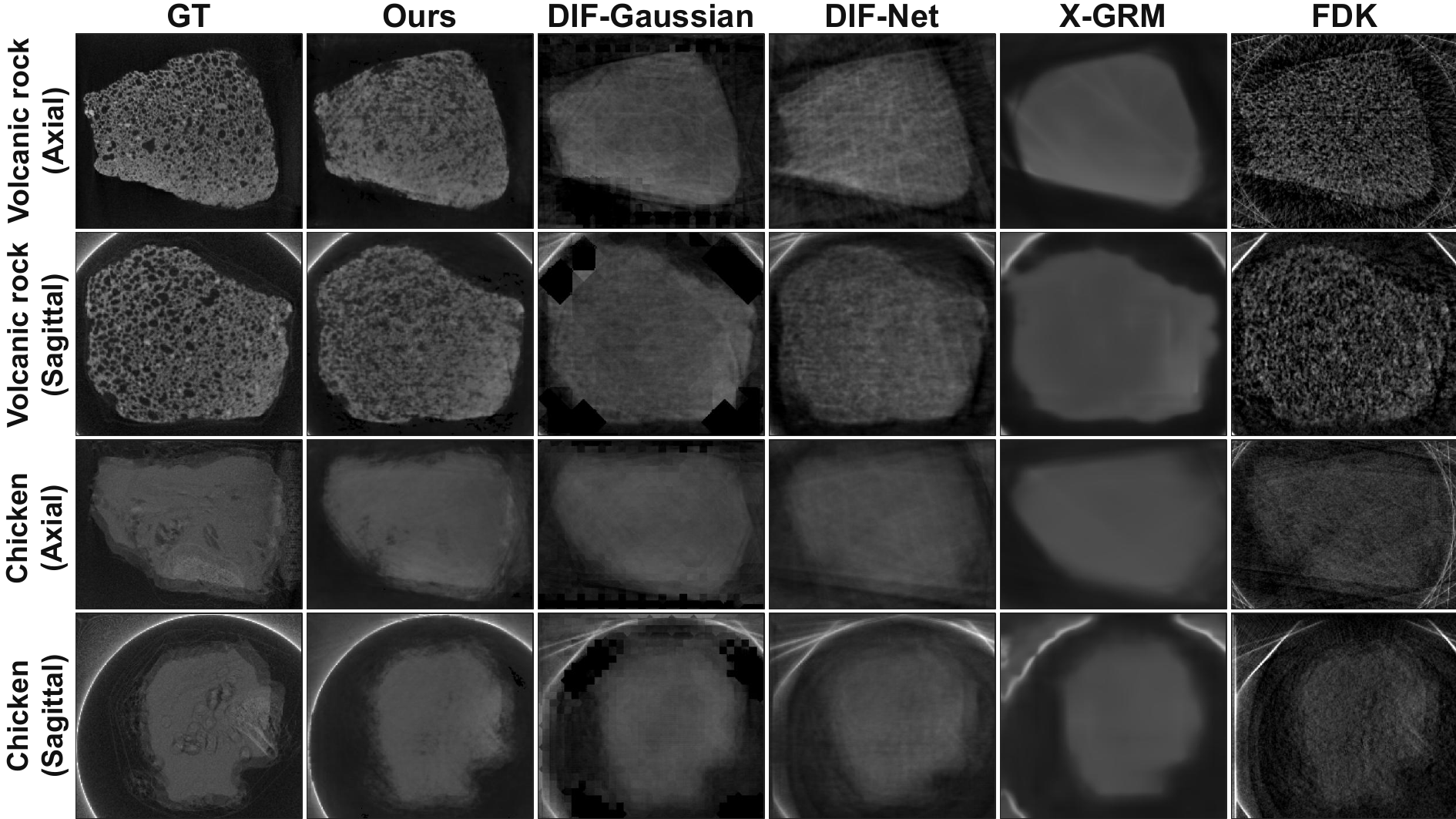}
\caption{Eighteen-view slices after adaptation. Columns are the reference, SparseTokenGS, DIF-Gaussian, DIF-Net, X-GRM, and FDK. Rows 1 and 2 are volcanic rock, axial slice 167 and sagittal slice 182. Rows 3 and 4 are chicken, axial slice 81 and sagittal slice 124.}
\label{fig:phantom}
\end{figure}

As Table~\ref{tab:phantom} and Fig.~\ref{fig:phantom} show, FDK reaches only $15.81$ dB and $16.09$ dB at eighteen views, because streaks from sparse angles are not corrected by a data-driven prior. X-GRM, DIF-Net, and DIF-Gaussian fix the appearance of the new object in local features, with PSNR of $19.79$ dB, $19.96$ dB, and $20.24$ dB on the rock, and $21.21$ dB, $21.18$ dB, and $21.50$ dB on the chicken. Later queries only interpolate these features, so the internal structure of the new object is hard to recover. SparseTokenGS still writes the projections into a fixed-length pool after adaptation, and the anchors and query coordinates continue to read that pool, so its PSNR reaches $25.71$ dB and $28.73$ dB, leading the next entry by $5.47$ dB and $7.23$ dB. On SSIM, X-GRM is next, at $0.5316$ and $0.5799$, and SparseTokenGS reaches $0.6802$ and $0.6861$. The slices are displayed over the full normalized intensity range $[0,1]$. Once the feed-forward baselines fix their local features, they no longer look back at the whole scan, which is why the margin on these new objects is larger.

\section{Discussion}
\label{sec:discussion}

\subsection{What Dual-Level Conditioning Adds}
Under sparse views one scan supplies few measurements, and if appearance is stored only in local anchors, a later coordinate query sees it after neighborhood interpolation, so structure supported by only a few views fades in the average. SparseTokenGS compresses the current projections into a fixed-length scan-level token pool and keeps it until decoding. The anchor read updates the local Gaussian features from the pool, and a Gaussian kernel then aggregates five neighboring anchors, so nearby positions share the same scan evidence and smoothly varying structures such as the chest wall, the lung boundary, and the dental arch form a continuous whole. The point read skips this aggregation and attends directly to the same pool, so a query can emphasize a token that the five-neighbor average has already discounted. Zero initialization lets the Gaussian branch follow the existing local estimate while the point branch reads the pool, so scan-level context enters in one forward pass. Removing the point-level read lowers LUNA16 by about $6$ dB in the ablation table, the largest drop among the rows. That drop shows the two reads are not substitutes: after aggregation, only the point read still sees the pool. Keeping both reads lets a smooth outline and a fine boundary be recovered together from the same sparse scan.

\subsection{Interpreting the Reconstruction Evidence}
In Table~\ref{tab:views}, SparseTokenGS attains the highest PSNR and SSIM among the compared entries on every dataset and at every view count. As more views are added, the encoder and the token pool receive fuller measurements, and the local features and the scan-level memory become richer together, so the scores rise accordingly. In Fig.~\ref{fig:luna_cor} the chest-wall and lung boundaries move closer to the reference, and Fig.~\ref{fig:four} preserves the broad anatomical outline on both the chest and dental cohorts. Once the feed-forward baselines write their anchors or voxels, that scan-level context stays fixed, so the added views cannot be read again at the query coordinate. Each view count uses its own checkpoint, while the pool length stays fixed. Additional views therefore enrich the same memory, and the advantage over the feed-forward baselines remains from three to eighteen views. The decoder and the anchor grid stay the same, so the gain comes from richer tokens in the same field.

The ablations show that these stages contribute together. The full checkpoint stands above the variants that remove anchor fusion, the token pool, or the similarity term. The largest gap appears when only the anchor read is kept: both LUNA16 and ToothFairy fall well below the main reference in Table~\ref{tab:ablation}. Neighborhood aggregation at the anchors forms the smoothly varying structure, and the point-level read supplies the scan-level evidence that this average does not deliver, so removing either one closes a distinct information path. Removing anchor fusion leaves the point-level read in place, and the score then falls by less than when the point-level read itself is removed. The similarity term holds neighboring samples together, so removing it costs more than removing only the anchor residual. In Fig.~\ref{fig:ablation}(b) the $14^3$ grid has the highest validation score: too few anchors leave the field insufficiently covered, while too many anchors make each anchor harder to fit, so $14^3$ is the operating point between coverage and learnability.

\subsection{Clinical Interpretation and Other Objects}
On the clinical dental scans, neighborhood aggregation at the anchors joins the dental arch and the mandible into a continuous outline, and the point-level read keeps the tooth and pulp boundary from depending only on that average. In Fig.~\ref{fig:real}, the six-view reconstruction stays consistent with the scanner's full-view reconstruction along these outlines. The larger absolute error still follows the tooth and cortical boundaries, where few angles leave those edges underdetermined, and the continuous arch remains available to the segmenter. In Fig.~\ref{fig:tips}, TIPs separates tooth and pulp without retraining. The same volumes reach $29.51$ dB against the scanner reconstruction, and the pulp Dice on the three displayed cases stays above $50\%$, so intensity agreement and segmentation jointly confirm the preservation of tooth and pulp.

The pool is rewritten from the current scan, while the encoder and the attention weights come from the training distribution. This same split between a smooth outline and a fine boundary also explains the results on other objects. On rock and chicken, a short same-object adaptation updates those weights, and the pool then organizes appearance according to the new object's reference. The advantage on both specimens in Table~\ref{tab:phantom} therefore comes from reusing this conditioning after the weights have seen the new intensity and geometry, with no redesign of the architecture. Feed-forward baselines lack a scan-level memory that remains readable at decoding, so once their local features are fixed they are hard to adjust.

\section{Conclusion}
\label{sec:conclusion}
In this work, we propose SparseTokenGS, a Gaussian-based method for sparse-view cone-beam computed tomography reconstruction via a reusable token pool. A pose-free pool is read at the anchors and at each query in one forward pass. This token-pool idea prepares a route for a family of feed-forward methods that retain a scan-level memory until coordinate decoding. Extensive experiments verify that it outperforms the compared feed-forward methods in image quality from three to eighteen views on three public datasets. Six-view clinical reconstruction supports tooth and pulp separation for lesion localization. In the future, we will develop it for rapid clinical lesion localization.

\section*{Acknowledgment}
All necessary clinical ethics approvals were obtained for the use of the in-house dental CBCT acquisitions and full-view reconstructions in this retrospective study. The Peking University Third Hospital Medical Science Research Ethics Committee approved the study under (2025) Medical Ethics Review No. (966-01), project IRB00006761-M20251203.

\providecommand{\refname}{References}

\end{document}